\documentclass[10pt,twocolumn,letterpaper]{article}

\usepackage[pagenumbers]{arxiv}

\usepackage[linesnumbered,ruled,vlined]{algorithm2e}
\usepackage{graphicx}
\usepackage{amsmath}
\usepackage{amssymb}
\usepackage{booktabs}
\usepackage{gensymb}
\usepackage{hyphenat}
\usepackage[inline]{enumitem}
\usepackage[breaklinks,colorlinks]{hyperref}
\usepackage{multirow}

\usepackage[capitalize]{cleveref}

\newcommand{\best}[1]{\textbf{\textcolor{red}{#1}}}
\newcommand{\sbest}[1]{\textbf{\textcolor{blue}{#1}}}

\begin{document}

\title{Post-Train Adaptive MobileNet for Fast Anti-Spoofing}

\author{Kostiantyn Khabarlak \\
Dnipro University of Technology\\
Ukraine\\
{\tt\small habarlack@gmail.com}
}

\maketitle

\begin{abstract}Many applications require high accuracy of neural networks as well as low latency and user data privacy guaranty. Face anti-spoofing is one of such tasks. However, a single model might not give the best results for different device performance categories, while training multiple models is time consuming. In this work we present Post-Train Adaptive (PTA) block. Such a block is simple in structure and offers a drop-in replacement for the MobileNetV2 Inverted Residual block. The PTA block has multiple branches with different computation costs. The branch to execute can be selected on-demand and at runtime; thus, offering different inference times and configuration capability for multiple device tiers. Crucially, the model is trained once and can be easily reconfigured after training, even directly on a mobile device. In addition, the proposed approach shows substantially better overall performance in comparison to the original MobileNetV2 as tested on CelebA-Spoof dataset. Different PTA block configurations are sampled at training time, which also decreases overall wall-clock time needed to train the model. While we present computational results for the anti-spoofing problem, the MobileNetV2 with PTA blocks is applicable to any problem solvable with convolutional neural networks, which makes the results presented practically significant.
\end{abstract}


\blfootnote{
  This article has been published as the peer-reviewed conference proceedings. Citation: \nohyphens{K.~Khabarlak, ``Post-Train Adaptive MobileNet for Fast Anti-Spoofing,'' in Proceedings of the 3rd International Workshop on Intelligent Information Technologies \& Systems of Information Security, Khmelnytskyi, Ukraine, March 23--25, 2022, vol.~3156, pp.~44–53. [Online]. Available: \url{http://ceur-ws.org/Vol-3156/keynote5.pdf}}
  
  The video presentation (keynote) is available at: \url{https://youtu.be/xmCCtd0Xj_0}
}

\section{Introduction}

Convolutional neural networks have shown extraordinary performance in computer vision tasks. While the initial research has been focused purely on quality regardless computational cost, the modern research trend is to design fast yet accurate neural networks. In significant part such a trend has been motivated by the requirements of low-latency data processing, user data privacy and reduction of server load. In addition to the fact that mobile and IoT devices offer significantly less computational power, typically several generations or price categories of such devices should be considered. Yet architecture of most modern neural networks can only be configured before training and not after. This leaves us with two alternatives:
\begin{enumerate*}[label={\arabic*)}]
  \item to train a separate network for each device category, which requires more time and effort;
  \item to design a single architecture which will target either high-end devices and high quality, or compatibility with all device generations at the cost of accuracy.
\end{enumerate*}
Both of these solutions are suboptimal.

The task of anti-spoofing is to distinguish whether the user shows its real, live face, or a recording of someone else's. Real-time face anti-spoofing is one of the algorithms that is preferable to be performed directly on a mobile device. The problem is complicated by a plethora of ways spoofing attack can be performed, such as printed face image, poster, video, face mask, \etc. Anti-spoofing can be found as a component in face-based access control systems. There it is not acceptable if access can be granted to an unauthorized person holding someone's photograph.

In this work we propose Post-Train Adaptive (PTA) block, which is simple in structure and offers a drop-in replacement for MobileNetV2 Inverted Residual block. The PTA block has multiple branches with different computation costs. The training procedure is constructed in way, so that in a fully-trained network the branch to infer on can be selected on-demand and at runtime. Thus, offering a way to change the network inference speed and to target multiple device tiers. Choice of block configuration can be made based on device speed, system load, desired power consumption or target quality. Crucially, the model is trained once and can be easily reconfigured after training, even directly on a mobile device.

To summarize, our main contributions are as follows:
\begin{enumerate}
  \item We introduce Post-Train Adaptive block for MobileNetV2 network, which is capable of switching between different performance/quality levels after being trained and even at runtime directly on a mobile device.
  \item We demonstrate superiority of the proposed approach over the original MobileNetV2 network both in terms of quality and inference speed on multiple mobile devices on face anti-spoofing problem. The qualitative metrics are provided on CelebA-Spoof dataset.
\end{enumerate}

\section{Literature Overview}
\subsection{Neural Network Architectures}\label{sec:neural-network-architectures}

The initial deep convolutional network research has been focused on finding exact configurations for convolution blocks (including kernel size and stride), pooling type and activation functions. Each block of these networks was ``plain'', \ie contained a single branch, such as in VGG network~\cite{VGG} with up to 19 layers deep. It has been noticed, that in general deep neural networks have better performance and overall generalization capability. However, it has turned out that building even deeper networks faces a vanishing gradient problem, when the training barely proceeds. In~\cite{ResNet} a training experiment has been conducted on plain networks with different depth. The first network contained 20 layers. The second network was constructed on top of the first one by adding more layers with the total of 56 layers. It has been expected, that if the newly added layers provide no additional benefit, they could be learned to produce an identity mapping. Hence, the deeper network should in theory show accuracy no less that the shallower network. Yet neural network training was very slow, which resulted in poor final accuracy. Further investigation has shown, that gradients computed in backpropagation algorithm were too small. This behavior is also known as vanishing gradient problem.

To counteract the vanishing gradient problem, the authors of ResNet network~\cite{ResNet} suggested adding an extra identity connection between groups of blocks. Such connection has been termed as skip or residual connection. The skip connection is used in the Bottleneck block, which is a group of 3 convolutions with kernel sizes of \(1 \times 1\), \(3 \times 3\), \(1 \times 1\) correspondingly. Skip connection is used around the block, so that the input to the first convolution is added to the result of the whole block. Besides, to limit the required computation in the block, the first \(1 \times 1\) convolution reduces the number of channels in the feature map, so that the heavier \(3 \times 3\) convolution processes smaller input. Finally, the last \(1 \times 1\) convolution increases back (restores) the number of channels. Hence, the name ``bottleneck''.

The authors of MobileNetV2~\cite{MobileNetV2} architecture improve on the ideas previously proposed in the ResNet network. MobileNetV2 is one of the most widely used networks on mobile devices. An Inverted Residual Block is introduced, which expands the number of channels inside the block while keeping few channels in the block inputs and outputs. From high level, this is an inversion of bottleneck block. 

Inverted residual block starts processing input with a \(1 \times 1\) convolution that expands the number of channels. The expansion is controlled by the expansion factor. The convolution is followed by the Batch Normalization~\cite{BatchNorm} and ReLU6 activation layer. Next, \(3 \times 3\) Depthwise Convolution is applied, followed again by the Batch Normalization and ReLU6. Lightweight depthwise convolution is used to decrease the number of computations. This type of convolution was not present in the original ResNet design. Finally, a \(1 \times 1\) convolution with Batch Normalization is applied to shrink the number of channel back to the original. Skip connection is used around the block. The proposed block architecture asymptotically reduces the number of multiply-add operations in the network, when compared to the bottleneck block. The MobileNetV2 network has a configurable width parameter, which tunes the network's computational complexity. The parameter is not adjustable after training is complete.

The following works have improved in the following directions: in Squeeze-and-Excitation Network (SENet)~\cite{SENet} an attention mechanism has been applied to improve the prediction quality. In neural networks, attention is used to selectively gate information flowing through the network, so that only the most important components of the signal flow forward. In MnasNet~\cite{MnasNet}, an approach for an automated neural architecture search for mobile and embedded devices has been proposed. During the neural network architecture selection process, the best network has been selected based on inference speed on an actual mobile device. MobileNetV3~\cite{MobileNetV3} has also improved on the previous approaches by using network architecture search, attention mechanisms and novel activation function. Large and small configurations have been introduced. Mobile neural network inference is important for many tasks, such as face processing~\cite{FastLandmark} or face anti-spoofing.

\subsection{Face Anti-Spoofing}

The goal of a face anti-spoofing system is to verify, that the face in front of the camera is a real face of a person standing and not a photograph being held to trick the system. Anti-spoofing is used to enhance camera-based access control systems to better detect unauthorized access. Depending on the access control system architecture, the verification process might need to be executed on a server, or directly on a mobile device (\eg~\cite{RfidAccessControl}).

Anti-spoofing can be performed based on RGB signal from a conventional camera, infrared or depth information from special hardware. For instance, CASIA-SURF~\cite{CasiaSurf} dataset has video information about all three modalities. Using them together improves overall quality, but depth or infrared information is typically not available and requires special hardware. Therefore, in this work we focus on algorithms that use RGB signal only. By using RGB signal anti-spoofing can still be performed by finding color and shape distortions. Neural networks show high quality in solving the anti-spoofing problem~\cite{CDC,CelebA-Spoof}, including the case when only RGB signal is present.

Neural-network-based anti-spoofing approaches are typically based on neural network architectures (backbones) described in \cref{sec:neural-network-architectures}. Several improvements have been proposed to these backbones. In~\cite{CDC} it was proposed to replace convolution operation with Central Difference Convolution, that better captures color gradients to improve anti-spoofing performance. In~\cite{CelebA-Spoof} AENet network was introduced with ResNet as a backbone. The authors utilize rich annotations of the CelebA-Spoof dataset (presented in the same work) to improve network training. Single multi-task loss is formed using face attribute information (\eg, smile, sunglasses, \etc), photo illumination conditions as well as depth and reflection information. Depth and reflection information is not inherently present in the dataset; hence, the authors propose to infer it from RGB image using an auxiliary neural network. The extra information is used during training and not required during inference.

\section{Materials and Methods}\label{sec:materials-methods}

While many of the above-described papers propose improvements to the neural network architectures, all of them require the architecture design to be completed prior to the training. No capability to change the network configuration after it has been trained is introduced, meaning that no single neural network can fully utilize hardware of devices with different performance characteristics.

\begin{figure*}
  \centering
  \includegraphics[width=0.7\linewidth]{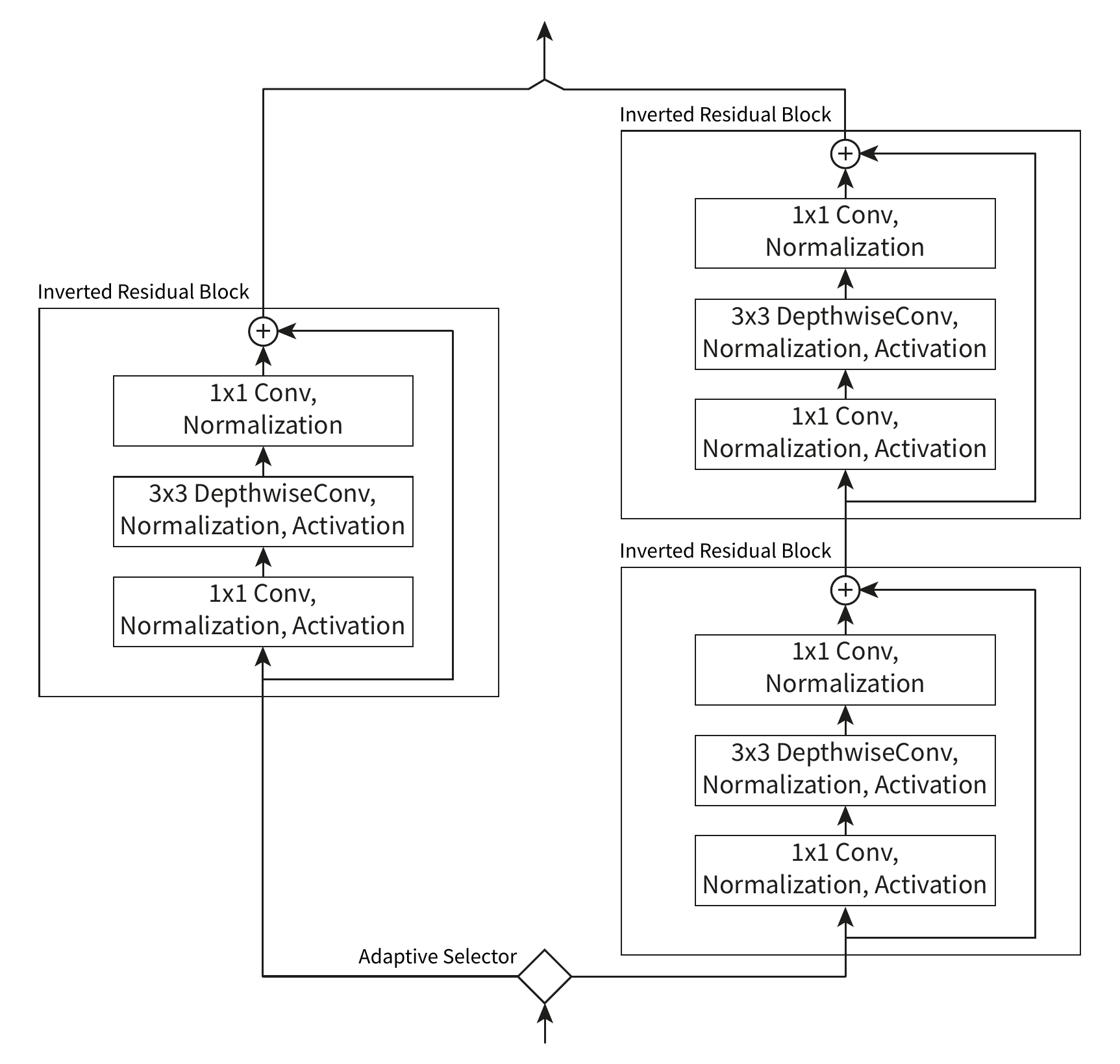}
  \caption{Post-Train Adaptive (PTA) block scheme. The block consists of 2 branches: Light (on the left), which contains a single Inverted Residual Block, and Heavy (on the right) with 2 Inverted Residual Blocks inferred sequentially. Adaptive Selector (at the bottom) can be configured to infer Light, Heavy, or Both branches. Branches have different inference time and target quality, which allows to better adapt the network to the target device or usage scenario. Crucially, the configuration can be done at runtime after the network training is complete.}\label{fig:pta-block}
\end{figure*}

To resolve the problem, in this work we propose a way to adaptively change network architecture based on user demand after the model training is complete. We base our approach on MobileNetV2 architecture as it is a widely-used mobile-friendly neural network. We introduce a Post-Train Adaptive (PTA) block as a base block for adaptive neural network inference. Its architecture is depicted in \cref{fig:pta-block}. The PTA block has 2 branches: the right (Heavy) branch is more computationally expensive and is fully equivalent to a pair of Inverted Residual Blocks; the left (Light) branch reduces the computation by executing only a single Inverted Residual Block. The branch to be executed is selected based on user configuration and can be changed dynamically at runtime. It is possible to execute either branch exclusively or both at the same time. If both branches are executed, their outputs are averaged element-wise, so that the feature distribution remains the same. The weights are not shared between any of the blocks. We propose to replace three pairs of Inverted Residual Blocks with the largest number of channels with PTA blocks in MobileNetV2 architecture, as is shown in \cref{tab:pta-architecture}.

\begin{table}
  \caption{Proposed MobileNetV2 with PTA Architecture. Block type, number of block output channels, number of block repeats, first block convolution stride are shown. In 3 places with the largest number of channels pairs of Inverted Residual blocks are replaced with Post-Train Adaptive (PTA) block.}\label{tab:pta-architecture}
  \centering
  \begin{tabular}{ccccc}
    \toprule
    Block Type        & Channels & Repeats & Stride \\
    \midrule
    Inverted Residual & 16       & 1       & 1       \\
    Inverted Residual & 24       & 2       & 2       \\
    Inverted Residual & 32       & 3       & 2       \\
    Inverted Residual & 64       & 2       & 2       \\
    \textbf{PTA}      & 64       & 1       & 1       \\
    Inverted Residual & 96       & 1       & 1       \\
    \textbf{PTA}      & 96       & 1       & 1       \\
    Inverted Residual & 160      & 1       & 2       \\
    \textbf{PTA}      & 160      & 1       & 1       \\
    Inverted Residual & 320      & 1       & 1       \\
    \bottomrule
  \end{tabular}
\end{table}

To train such a model we randomly sample a configuration of PTA blocks at each iteration, and perform forward, then backward pass updating the weights. To avoid excessive randomness in the model, we limit the number of possible configurations for the model to 5:
\begin{itemize}
  \item all blocks execute the heavy branch;
  \item a single of the blocks executes the light branch, while others the heavy one (3 possible combinations);
  \item all of the blocks execute the light branch.
\end{itemize}
Configuration sampling is not performed uniformly, we follow the intuition that paths with larger number of weights should be trained for longer and therefore assign higher sampling probabilities to such configurations. The exact sampling probabilities are shown in \cref{tab:pta-sampling}. Note, that we do not execute both branches at the same time during training. All configurations missing from \cref{tab:pta-sampling} are also assumed to be never sampled and trained on.

\begin{table}
  \caption{PTA configurations and corresponding sampling probabilities during training time. Configurations with more weights are assigned higher sampling probabilities. Note that both branches are never executed at the same time during training.}\label{tab:pta-sampling}
  \centering
  \begin{tabular}{lc}
    \toprule
    PTA Configuration     & Sampling Probability \\
    % \relax + [...] prevents what's inside brackets from being evaluated as command parameter
    \midrule \relax
    [Heavy, Heavy, Heavy] & 0.45                 \\ \relax
    [Light, Heavy, Heavy] & 0.15                 \\ \relax
    [Heavy, Light, Heavy] & 0.15                 \\ \relax
    [Heavy, Heavy, Light] & 0.15                 \\ \relax
    [Light, Light, Light] & 0.10                 \\ \relax
    [Both, Both, Both]    & 0.00                 \\
    \bottomrule
  \end{tabular}
\end{table}

Anti-spoofing is a classification problem, so to train the model we use Cross Entropy as a loss function. The model that we consider output logits, therefore, Cross Entropy also includes softmax computation, and is defined as follows:
\pagebreak
\begin{equation}
  l(x, y) = -\sum_{n=1}^N{\sum_{c=1}^C{\log{ \frac{\exp{(x_{n,c})}}{\exp{(\sum_{i=1}^C{x_{n,i}})}}}}}
\end{equation}
where \(N\) is the number of samples in a mini-batch, \(C = 2\) is the number of classes, \(x_{n,c}\) is the model logit output for item \(n\) and class \(c\).

\section{Experiments}

To train and evaluate the model we use recent CelebA-Spoof~\cite{CelebA-Spoof} dataset, following~\cite{CelebA-Spoof,CelebA-Spoof-Challenge}. To the best of our knowledge, this is the largest anti-spoofing dataset available to date. Overall, it contains 625,537 pictures (including both spoof and live photos) of 10,177 subjects. Photos are captured with different lighting, environment conditions and different cameras. Only RGB photos are available in the dataset. Several spoof attack types are considered in the dataset, such as printed full frame photos, paper cut photos, replay attack when picture is presented on a tablet or a phone, and a 3D mask when a printed image is overlayed on top of a human face. In addition to binary spoof/non-spoof label, the dataset contains rich information about spoof type, illumination condition, environment label as well as face attribute labels (smile, mustache, hat, eyeglasses, \etc). At this point we use only binary spoof/non-spoof information in our model with a possibility of extending it in the future. The dataset defines train/test splits and several evaluation protocols. Our results are given for the intra-test protocol, which is used for general model evaluation. We split the training subset randomly into actually training and validation in 80/20 ratio. Training is performed for 20 epochs. The best model is then selected based on validation set. Gradient computation is performed on mini-batches of 32 images. Adam~\cite{Adam} adaptive gradient descent method with learning rate \(\alpha = 10^{-4}\) is used as an optimizer. The results are reported on the test set.

Images are cropped based on face bounding boxes that are present for each image in the CelebA-Spoof dataset. The resulting face image is then resized to the resolution of \(128 \times 128\). We feed color (RGB) images to the model. At training time color jitter and ISO-noise augmentations are used. Note, that no ImageNet pretraining has been used, the models are trained from scratch.

We follow~\cite{DualSpoof,BiometricPresentationAttack} and use the following metrics for model quality evaluation in our paper:
\begin{itemize}
  \item Accuracy is a proportion of correctly classified images to the overall number of images;
  \item Attack Presentation Classification Error Rate (APCER) is a proportion of attack images incorrectly classified as normal images:
  \begin{equation}
    APCER = \frac{FP}{FP + TN};
  \end{equation}
  \item Bona Fide Presentation Classification Error Rate (BPCER), that is a proportion of normal (bona) images incorrectly classified as attack images:
  \begin{equation}
    BPCER = \frac{FN}{FN + TP};
  \end{equation}
  \item Average Classification Error Rate (ACER) is an average of APCER and BPCER:
  \begin{equation}
    ACER = \frac{APCER + BPCER}{2},
  \end{equation}
\end{itemize}
where TP is True Positive, that is a sample is labelled as spoof and the prediction is correct; TN is True Negative, meaning both prediction and true label are non-spoof; FP is False Positive, \ie, prediction is spoof, while the image is non-spoof; finally, FN is False Negative, the prediction is non-spoof, but actual image is spoofed.

As in this paper we not only target adaptivity and quality, but also inference time on a mobile device, we have selected a pair of Android smartphones for testing. The devices are based on Qualcomm Snapdragon 845 and Snapdragon 800 CPUs, the flagship mobile processors from 2018 and 2013 correspondingly. In terms of modern-day processors, the former can be though-of as a mid-to-high-end CPU, and the latter as a low-end CPU. These processors are found in many devices; there, our results can be easily reproduced.

In addition, we report training time for both MobileNetV2 and MobileNetV2 with PTA blocks on GTX 1050Ti GPU, which is an important metric for practical applications.

\section{Results}

For the comparison 2 models have been trained: the original MobileNetV2 (hereinafter No PTA) and MobileNetV2 with PTA blocks (hereinafter PTA), constructed as described in \cref{sec:materials-methods}. PTA-based models can be further configured after being training, so in all of the following tables we show the configuration for which the testing has been performed. As the proposed MobileNetV2+PTA configuration consists of 3 PTA blocks we use 3-letter abbreviation to denote the exact configuration used. Letters H, L, B are used to denote the execution of Heavy, Light, and Both branches correspondingly for each of the PTA blocks.

In \cref{tab:pta-quality} shows qualitative results as measured on the test set. Accuracy is the higher the better. APCER, BPCER, ACER denote error rates, consequently, the lower the better. The best result in each column is shown in red, the second best is shown in blue. As can be seen, PTA-based models dominate the original MobileNetV2 (No PTA) implementation in all of the metrics. Interestingly, PTA-HHH configuration, which is equivalent in terms of number of parameters and multiply-additions is also better than the original model.

\begin{table*}
  \caption{Model quality comparison of the original MobileNetV2 (no PTA) vs PTA-based configurations (our work). All PTA configurations correspond to the same model, trained once. Accuracy as well as anti-spoofing error metrics (APCER, BPCER, ACER) are used for the comparison. The best result is shown in red, the second best is shown in blue. As is clearly seen, PTA-based models dominate the non-PTA version in every metric.}\label{tab:pta-quality}
  \centering
  \begin{tabular}{lcccc}
    \toprule
    Configuration & Accuracy (\(\uparrow\), \%) & APCER (\(\downarrow\), \%) & BPCER (\(\downarrow\), \%) & ACER (\(\downarrow\), \%) \\
    \midrule
    No PTA        & 96.74                       & 1.07                       & 4.18                       & 2.63                      \\
    PTA-HHH       & 96.89                       & \sbest{0.72}               & 4.12                       & 2.42                      \\
    PTA-LHH       & 96.84                       & \best{0.70}                & 4.20                       & 2.45                      \\
    PTA-HLH       & 97.04                       & 0.80                       & 3.88                       & 2.34                      \\
    PTA-HHL       & \sbest{97.83}               & 2.30                       & \sbest{2.11}               & \sbest{2.21}              \\
    PTA-LLL       & \best{97.85}                & 2.53                       & \best{1.98}                & 2.26                      \\
    PTA-BBB       & 97.49                       & 1.21                       & 3.05                       & \best{2.13}               \\
    \bottomrule
  \end{tabular}
\end{table*}

In Table~\ref{tab:pta-inference-time} we present model complexity and inference time comparison. First, we show the number of parameters in each of the models (in millions). For the PTA models we configure the model after training, and then report the number of parameters that is actively used in the corresponding configuration. Next, we measure the number multiply-add operations (in million operations) executed during the forward pass of the model. Also, we measure actual performance on mobile devices on widely popular Snapdragon 845 and Snapdragon 800 processors (SD845 and SD800 correspondingly), measured in milliseconds. Finally, we show relative inference time improvement with respect to the No PTA baseline as measured on SD845. PTA-HHH configuration has the same number of parameters and computation as No PTA and is equivalent in terms of performance on a real device. PTA-BBB configuration uses both Light and Heavy branches in all 3 PTA blocks and therefore is slightly more computationally intensive. All other configurations that use a mix of Heavy and Light blocks are faster than MobileNetV2.

\begin{table*}
  \caption{Model complexity and inference time comparison of MobileNetV2 (No PTA) vs PTA-based configurations (this work). Number of model parameters, estimated number of multiplication and addition operations, inference time on Snapdragon 845 and 800 mobile processors, and relative inference speedup are shown. The best result is shown in red, the second best is in blue. Model inference speed can be configured on demand from 107\% to 80\% relative to the MobileNetV2 baseline. PTA-LLL is the lightest model. PTA-HHH is equivalent in model architecture and is as fast as the original MobileNetV2, meaning PTA-based architecture does not introduce any overhead.}\label{tab:pta-inference-time}
  \centering
  \begin{tabular}{lccccc}
    \toprule
    Configuration & \# Params           & Multiply-Adds          & Inference Time             & Inference Time             & Relative Inference    \\
                  & (\(\downarrow\), M) & (\(\downarrow\), Mops) & SD845 (\(\downarrow\), ms) & SD800 (\(\downarrow\), ms) & Time (\(\downarrow\)) \\
    \midrule
    No PTA        & 2.23                & 104.15                 & 21.32                      & 94.23                      & 1.00                  \\
    PTA-HHH       & 2.23                & 104.15                 & 21.27                      & 94.12                      & 1.00                  \\
    PTA-LHH       & 2.17                & 100.63                 & \sbest{18.61}              & \sbest{82.24}              & \sbest{0.87}          \\
    PTA-HLH       & 2.11                & \sbest{96.50}          & 19.67                      & 86.96                      & 0.92                  \\
    PTA-HHL       & \sbest{1.91}        & 99.00                  & 19.27                      & 85.15                      & 0.90                  \\
    PTA-LLL       & \best{1.73}         & \best{87.84}           & \best{17.08}               & \best{75.44}               & \best{0.80}           \\
    PTA-BBB       & 2.73                & 120.47                 & 22.72                      & 100.47                     & 1.07                  \\
    \bottomrule
  \end{tabular}
\end{table*}

As we sample Light and Heavy PTA configurations during training, it is expected that overall training time for the PTA-based model should decrease. Our experiments validate this assumption. In Table~\ref{tab:pta-training-time}, we demonstrate training time for each of the models joined by the best Accuracy and ACER achieved by each of the models. We show epoch training time in minutes, and overall training time for 20 epochs in hours. As can be seen, PTA-based model is better in terms of quality, inference and training time.

\begin{table*}
  \caption{Training time, Accuracy and ACER comparison for the original MobileNetV2 vs MobileNetV2 with PTA blocks. The best result is shown in red. PTA-based train-time configuration sampling decreases overall training time, while improving resulting model quality.}\label{tab:pta-training-time}
  \centering
  \begin{tabular}{lcccc}
    \toprule
    Configuration   & Epoch Training           & Overall Training         & Best Accuracy      & Best ACER            \\
                    & Time (\(\downarrow\), m) & Time (\(\downarrow\), h) & (\(\uparrow\), \%) & (\(\downarrow\), \%) \\
    \midrule
    MobileNetV2     & 49.28                    & 16.43                    & 96.74              & 2.63                 \\
    MobileNetV2+PTA & \best{43.11}             & \best{14.37}             & \best{97.85}       & \best{2.13}          \\
    \bottomrule
  \end{tabular}
\end{table*}

\begin{figure}
  \centering
  \includegraphics[width=\linewidth]{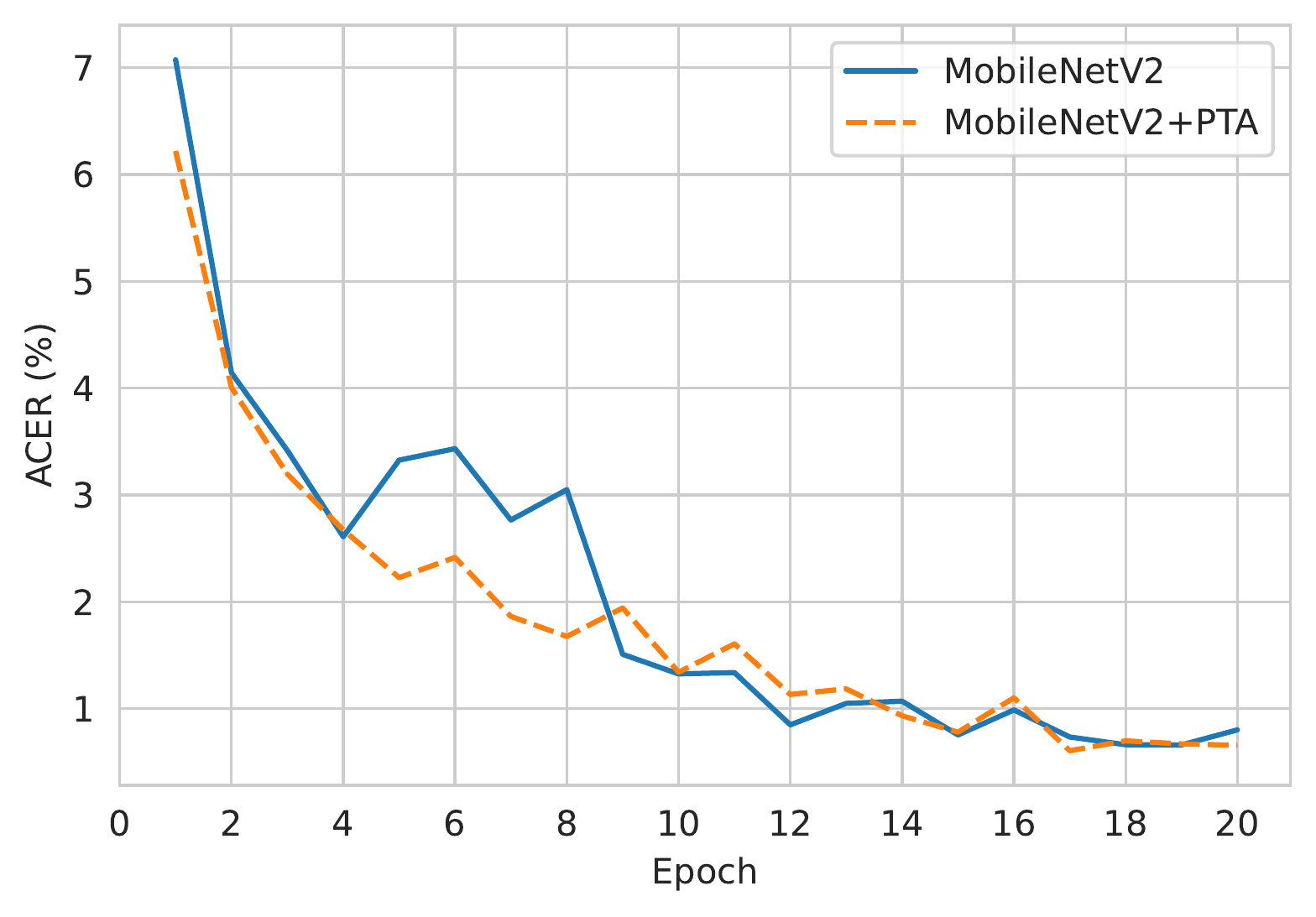}
  \caption{Validation ACER (the lower, the better) for MobileNetV2 (solid blue line) vs MobileNetV2 with PTA blocks (dashed orange line). The metric is measured on PTA-HHH configuration. As is clearly visible, with PTA-based model ACER decreases faster.}\label{fig:pta-training-acer}
\end{figure}

In Figure~\ref{fig:pta-training-acer} we show validation ACER during training for the original MobileNetV2 (solid blue line) and MobileNetV2+PTA (dashed orange line). For the PTA-based model we validate on PTA-HHH configuration, which is equivalent in terms of the number of parameters and multiply-adds to the original MobileNetV2. As is clearly seen, ACER error rate decreases faster with MobileNetV2+PTA than with the original MobileNetV2.

\section{Discussion}

The key goal of this work is to make it possible to reconfigure a neural network after it has been trained. As has been shown, Post-Train Adaptive block proposed in this work is an efficient way for post-train network configuration. Placing only 3 PTA blocks in the MobileNetV2 has made it possible to adaptively adjust inference time from 107\% to 80\% of the original MobileNetV2 (see \cref{tab:pta-inference-time}). The simplicity of the PTA block has allowed to implement MobileNetV2 with PTA for inference on mobile devices. Different CPUs were tested: high-end Snapdragon 845 and low-end Snapdragon 800. On the latter inference speed improvement over the original MobileNetV2 is over 18 milliseconds, which is significant for a performance-limited device. Interestingly, the second best inference time is achieved by the PTA-LHH configuration with 100.63~Mops and not by PTA-HLH with 96.50~Mops. PTA-LHH is slower than PTA-HLH in 4.72~ms on SD800, while both of them are faster when compared to the No PTA model. As expected, the best inference speed is offered by the PTA-LLL configuration, where all three PTA blocks use the light branch only.

PTA-based configurations have better quality than the original MobileNetV2 as well. MobileNetV2 (No PTA) and PTA-HHH configurations have the same number of parameters and multiply-adds, but PTA-HHH is better in every metric as seen from \cref{tab:pta-quality}. In this case, the only difference is in the newly-proposed training procedure with PTA configuration sampling. Consequently, we suggest, that the proposed training procedure has a positive impact on overall model quality.

Validation ACER comparison depicted in \cref{fig:pta-training-acer} shows that PTA-HHH is better than No PTA model throughout the training procedure. For instance, after a single training epoch these models have achieved ACER of 6.46\% (PTA-HHH) and 9.3\% (No PTA), meaning PTA-based model starts to train significantly faster. The final validation ACER for PTA-HHH is 0.53\% and is 1.0\% for No PTA. We suggest that sampling different block configurations during training makes the network learn more general features and offers regularization capability. This might explain better PTA-based model results.

When comparing all PTA-based configurations, the best Accuracy and BPCER is shown by PTA-LLL at 97.85\% and 1.98\% correspondingly. This is also the fastest configuration. PTA-LHH has the lowest APCER at 0.70\% (53\% relative improvement to the MobileNetV2).

We also investigate a possibility of using multiple branches jointly in PTA-BBB configuration. The 2-branch PTA-BBB model has more parameters (2.73~M) and multiply-additions (120.47~Mops) than the original MobileNetV2 model with 2.23~M and 104.15~Mops correspondingly, and is only slightly slower. This is the configuration that shows the best performance in average classification error rate (ACER) at 2.13\%, which is a 23.5\% relative improvement over the No PTA baseline. The model is also better than MobileNetV2 (No PTA) in all other metrics. Note, that the PTA-based network is never actually trained with both branches enabled. Therefore, we expect each of the branches to learn slightly different features; thus, forming as in-model ensemble similar to Dropout~\cite{Dropout} technique.

Overall PTA-based model training time is lower than that of the conventional MobileNetV2 as can be seen from \cref{tab:pta-training-time}. On a mainstream Nvidia GTX 1050Ti graphics card we see 2-hour (or 14\%) overall training time reduction, when the model is trained for 20 epochs. This significant model training time decrease is achieved by two facts:
\begin{enumerate*}[label={\arabic*)}]
  \item the heaviest PTA-BBB configuration is never used during training, as is previously mentioned;
  \item the actual configurations sampled during training (see \cref{tab:pta-sampling}) are on average lighter than MobileNetV2 (No PTA), which results in fewer multiply-add operations performed during both inference and training time.
\end{enumerate*}
Note, all PTA configurations are obtained from a single trained model. This differs our approach from other found in literature.

\section{Conclusion}

In this work Post-Train Adaptive block has been first introduced. Such a block is simple in structure and offers a drop-in replacement for a pair of MobileNetV2 Inverted Residual blocks. Thanks to the proposed novel block we improve over MobileNetV2 for anti-spoofing in the following ways:
\begin{enumerate}
  \item We solve the problem of inability to change the network architecture after it has been trained. The PTA block has Light and Heavy branches. The branch to execute can be selected on-demand and at runtime, even on a mobile device. Not only each of the branches can be used exclusively, but also their prediction can be averaged, forming an in-model ensemble. Therefore, a model can be reconfigured after training to better suit the target device.
  \item The lightest PTA configuration shows 20\% improvement in terms of actual inference speed on a mobile device, while also having superior quality in comparison to the original MobileNetV2 architecture.
  \item The anti-spoofing performance has been substantially improved with PTA-based configurations beating the baseline in all typical anti-spoofing metrics.
\end{enumerate}

During training we sample PTA configurations with different number of parameters. We suggest that this results in the model learning more general features and in better overall quality. All of the aforementioned improvements have been achieved with smaller total training time in comparison to the MobileNetV2 model.

Because-of a significant variation of mobile and edge device computational power, a single neural network targeting several different device categories is suboptimal. The proposed approach allows to train the model once and then adjust its inference speed according to device characteristics, overall system load and desired battery consumption. This makes the results obtained practically significant.

In this work we have investigated only a single (yet important) practical application of the MobileNetV2 with PTA blocks, that is face anti-spoofing. However, the proposed PTA blocks are applicable to any problem, that can be solved using convolutional neural networks. In future works we will expand our exploration on other applications and will improve PTA block performance and quality even further.

\begin{small}
  \subsection*{Funding}

  The work is supported by the state budget scientific research project of Dnipro University of Technology ``Development of New Mobile Information Technologies for Person Identification and Object Classification in the Surrounding Environment'' (state registration number 0121U109787).

  \printbibliography

\end{small}

\end{document}